\pdfoutput=1
\documentclass[11pt]{article}

\usepackage[table,xcdraw]{xcolor}
\usepackage[final]{acl}

\usepackage{times}
\usepackage{latexsym}
\usepackage[T1]{fontenc}
\usepackage[utf8]{inputenc}

\usepackage{microtype}

\usepackage{inconsolata}

\usepackage{graphicx}

\title{Training LLMs to Recognize Hedges in Spontaneous Narratives}
\usepackage{todonotes}

\newcommand{\repo}{https://github.com/cogstates/hedging}

\usepackage{booktabs}
\usepackage{multirow}

\usepackage{mdframed}
\usepackage{listings}
\lstdefinestyle{minted}{
    basicstyle=\footnotesize\ttfamily\linespread{4},
    breaklines=true,
    columns=flexible,
    commentstyle=\color[rgb]{0.127,0.427,0.514}\ttfamily\itshape,
    identifierstyle=\color{black},
    inputencoding=latin1,
    keywordstyle=\color[HTML]{228B22}\bfseries,
    ndkeywordstyle=\color[HTML]{228B22}\bfseries,
    prebreak=\raisebox{0ex}[0ex][0ex]{\ensuremath{\hookleftarrow}},
    showstringspaces=true,
    stringstyle=\color[rgb]{0.639,0.082,0.082}\ttfamily,
    upquote=true
}
\surroundwithmdframed[
    backgroundcolor=gray!10, linecolor=gray!60,
    innertopmargin=0, innerbottommargin=0
]{lstlisting}

\usepackage{glossaries}
\makenoidxglossaries
\renewcommand{\glossarysection}[2][]{}

\newglossaryentry{narrative element}
{
    name=narrative element,
    description={Observable events in the Roadrunner cartoon that and were likely to be mentioned in narrations \cite[see][]{GalatiBrennan2010}. Segmentation by narrative elements allowed for comparisons across speakers for elements realized in each narration}
}
\newglossaryentry{Cohen's Kappa}
{
    name=Cohen's Kappa,
    description={Measure of agreement between two raters that an item falls within a subjective category; higher values denote higher agreement}
}
\newglossaryentry{LSTM}
{
    name=LSTM,
    description={Long Short-Term Memory networks \citep{hochreiter-1997-long} are a type of recurrent neural network designed to capture long-range dependencies}
}
\newglossaryentry{BERT}
{
    name=BERT,
    description={BERT \citep{devlin-etal-2018-bert} stands for Bidirectional Encoder Representations from Transformers.  BERT is a transformer-based model which produces contextual representations of text by conditioning on both the left and right surrounding words}
}
\newglossaryentry{LLM}
{
    name=LLM,
    description={Large Language Models are large (typically by parameter count) models which take in text and produce a distribution over their vocabulary which can be used to predict the next token},
}
\newglossaryentry{pos-tag}
{
    name=part-of-speech tag,
    description={TODO}
}
\newglossaryentry{cross-validation}
{
    name=cross-validation,
    description={TODO}
}
\newglossaryentry{few-shot}
{
    name=few-shot,
    description={TODO}
}
\newglossaryentry{F1}
{
    name=F1,
    description={
    The harmonic mean of \gls{precision} and \gls{recall}. $$F1 = 2\cdot\frac{\text{precision} \cdot \text{recall}}{\text{precision} + \text{recall}}$$ It is also called F-measure or F-score. 
    Loosely speaking, the metric is a balance of how often the model is correct when it predicts a particular class (precision), and how often the model predicts that class when it would be correct to do so (recall)}
}
\newglossaryentry{precision}
{
    name=precision,
    description={The number of correct predictions (true positives) for a class divided by the number of times the model predicted that class (true positives + false positives)}
}
\newglossaryentry{recall}
{
    name=recall,
    description={The number of correct predictions (true positives) for a class divided by the number of samples which belong to that class (true positives + false negatives)}
}
\newglossaryentry{BIO}
{
    name=BIO,
    description={BIO, short for Beginning, Inside, Outside, is a format for labeling chunks of tokens. Tokens are assigned B if they begin a sequence which should be labeled (e.g., a named entity), I if they belong to a previously begun sequence, and O otherwise}
}
\newglossaryentry{temperature}
{
    name=temperature,
    description={A hyperparameter that modifies the next token distribution of language models. Larger temperature values increase the likelihood of lower probability tokens}
}
\newglossaryentry{token}
{
    name=token,
    description={The smallest unit of text, often words or subwords, which are used as the input for various NLP models}
}
\newglossaryentry{epoch}
{
    name=epoch,
    description={A single pass through the training data}
}

\newcommand{\tss}[1]{\textsuperscript{#1}}

\usepackage[pro]{fontawesome5}
\usepackage{fdsymbol}
\newcommand{\DoP}{{\(\Psi\)}}
\newcommand{\DoL}{{\tiny\faComments}}
\newcommand{\DoCS}{{\tiny\faLaptop}}
\newcommand{\IACS}{{\tiny\faCalculator}}
\author{
    Amie J. Paige\tss{$\ast$\hspace{0.1em}\DoP}, 
    Adil Soubki\tss{$\ast$\DoCS\hspace{0.1em}\IACS}, 
    John Murzaku\tss{$\ast$\DoCS\hspace{0.1em}\IACS}, 
    Owen Rambow\tss{\DoL\hspace{0.15em}\IACS}, 
    Susan E. Brennan\tss{\DoP} \\
    \tss{\DoCS}Department of Computer Science,
    \tss{\DoL}Department of Linguistics,
    \tss{\DoP}Department of Psychology \\
    \tss{\IACS}Institute for Advanced Computational Science,
    Stony Brook University  \\
    \tss{$\ast$}These authors contributed equally to this study. \\
    \texttt{amie.paige@stonybrook.edu,}
    \texttt{\{asoubki,jmurzaku\}@cs.stonybrook.edu}
}

\begin{document}
\maketitle

%JM 05/20 - fixed Owen's PDF comments on abstract
%\john{Addressed Owen's PDFs comments for abstracr. @Susan @Amie lmk if you like the rewrite and fixes.}
\begin{abstract}
Hedges allow speakers to mark utterances as provisional, whether to signal non-prototypicality or “fuzziness”, to indicate a lack of commitment to an utterance, to attribute responsibility for a statement to someone else, to invite input from a partner, or to soften critical feedback in the service of face-management needs. Here we focus on hedges in an experimentally parameterized corpus of 63 Roadrunner cartoon narratives spontaneously produced from memory by 21 speakers for co-present addressees, transcribed to text \citep{GalatiBrennan2010}.
We created a gold standard of hedges annotated by human coders (the \textit{Roadrunner-Hedge corpus}) and compared three LLM-based approaches for hedge detection: fine-tuning BERT, and zero and few-shot prompting with GPT-4o and LLaMA-3. The best-performing approach was a fine-tuned BERT model, followed by few-shot GPT-4o. After an error analysis on the top performing approaches, we used an \textit{LLM-in-the-Loop} approach to improve the gold standard coding, as well as to highlight cases in which hedges are ambiguous in linguistically interesting ways that will guide future research. This is the first step in our research program to train LLMs to interpret and generate collateral signals appropriately and meaningfully in conversation.
\end{abstract}

\section{Introduction}
%\john{5/20Addressed owen's comments here. }
% \blfootnote{\tss{$\ast$}Denotes equal contribution to this paper.}
The virtuosity of \gls{LLM}s such as ChatGPT has led some to the impression that AI already converses (or will soon be able to converse) as people do. But as language users, LLMs and humans are quite different. The underlying foundations for learning by these distinct kinds of language users share little in common: Humans learn as infants to interact with others well before they learn their first words, and once word learning begins, they can pick up a new word in one or just a few exposures, whereas LLMs are pre-trained on humanly unfathomable quantities of text without ever learning to interact. Transformer-based chat programs can generate paragraphs-worth of text remarkably well without modeling the coordination between agents--but is this conversation?

Whether a sequence of prompts and responses exchanged in a dialogue between an LLM agent and a human counts as truly (rather than superficially) ``conversational'' depends on how conversation is conceptualized. Conversation is often presumed to be the passing back and forth of messages (a ``message model''); but that does not explain phenomena common to spontaneous conversation such as incremental turns, clarifications, and repair. Here we conceptualize conversation as a collaborative process of grounding meanings (seeking and providing evidence) during which two or more partners signal, coordinate, and align their beliefs or cognitive states \citep{brennan2005conversation, CLARK1986}. This leads to a broader research agenda that we hope will push generative AI to model phenomena such as a partner’s knowledge or theory of mind, mutual beliefs or common ground, as well as when to take initiative in a dialogue.

The main contributions of this work include:

\noindent (i) After grounding the project in psycholinguistic theory (Section~\ref{sec:theory}) and related work (Section~\ref{sec:rel_work}), we present the Roadrunner-Hedge Corpus (Section~\ref{sec:corpus}), a corpus of spontaneous face-to-face narratives annotated for hedging.\footnote{\url{\repo}}

\noindent (ii) We describe a set of experiments on this corpus using zero-shot, few-shot, and fine-tuning methods on modern LLMs (Section~\ref{sec:experiments}).

\noindent (iii) We perform a detailed error analysis pinpointing where LLMs fail in detecting hedges (Section~\ref{sec:ea}). With this analysis, we take an LLM-in-the-Loop approach to correcting gold annotations, reducing errors in our top performing systems.

% This paper is organized as follows. After summarizing psycholinguistic theory relevant to hedging in Section~\ref{sec:theory} and related work in Section~\ref{sec:rel_work}, we describe the Roadrunner corpus Section~\ref{sec:corpus}. The experiments are described in Section~\ref{sec:experiments} and provide a error analysis in Section~\ref{sec:ea}. We conclude with a discussion and implications of our results in Section~\ref{sec:discussion}, limitations and the future of our work in \ref{sec:fw}, and a final summary of our salient contributions in Section~\ref{sec:conclusion}.
We conclude with a discussion and implications of our results in Section~\ref{sec:discussion}, limitations and the future of our work in Section \ref{sec:fw}, and a final summary of our salient contributions in Section~\ref{sec:conclusion}.

\section{Theoretical Foundations from Psycholinguistics}
\label{sec:theory}
In conversation, people communicate not only about the purpose or topic at hand, but they also communicate meta-information about what they're saying within the context of interaction, or \textit{collateral signals} \cite{Clark_1996}. Along with providing evidence for grounding in conversation, about whether a prior turn has been understood as intended \cite{clark1991grounding}, collateral signals can also provide information about the speaker’s relationship with the content of their message—how confident they are in what they are saying, whether it is difficult to recall or express, and whether they would welcome input from their partner. In this project, we focus on a particular kind of collateral signal used for coordination, \textit{hedges}.

\subsection{Why Speakers Hedge}

There have been several proposals for why speakers hedge. Hedges have been claimed to characterize powerless “feminine” language \cite{RLakoff1973} or to serve a politeness function by minimizing threat to a partner’s “face” \citep{BrownLevinson1987}; see also \cite{Fraser}. Hedges have also been thought to convey a certain “fuzziness” of category membership when a speaker means to describe a non-prototypical member of a category \citep[e.g., a penguin belonging to the bird category;][]{Lakoff1975}.
%\adil{unmatched parens here?} 
%Prince and colleagues (1982) 
\citet{Prince1982}
suggested that hedges play two functions: First, to make propositional content less exact (\textit{approximators}, e.g. “sort of”) and second, to change the relationship a speaker has to the content of their message (\textit{shield hedges}). Shield hedges are further divided into \textit{plausibility} shields that signal a lack of commitment to the content of a message \cite[``I think his feet were blue,''][p. 5]{Prince1982}, and \textit{attribution} shields that assign responsibility for a message to a source other than the speaker or writer themself \cite[``According to her estimates...''][p. 13]{Prince1982}.

Several experimental studies have demonstrated how hedges can convey speakers' commitment to what they are saying.
For example, in a question-answering task, people trying to recall the answers to trivia questions produced more disfluencies, longer latencies, more rising intonation, and more expressions of doubt when they reported having a low \textit{feeling of knowing} about an answer. This metacognitive information was confirmed to be accurate when compared to the ground truth in the form of their answer to the same (multiple-choice) question later \cite{SMITH1993}. Not only are hedges informative as collateral signals about what a speaker knows, but they are accurately interpreted as such by listeners \cite{brennan1995feeling}. 

That hedges function as interactional signals in extended dialogue is evident from studies of referential communication. Typically in such studies, two partners who can’t see each other converse in order to arrange and rearrange duplicate sets of objects in matching orders, with the objects needing to be distinguished from similar objects or consisting of Tangrams (abstract geometric shapes unassociated with any conventional or lexicalized labels). Hedges are common in initial referring expressions, where they tend to appear in wordy, disfluent, and often tentative descriptions, and then they drop out in repeated referring expressions once partners have reached a shared conceptualization for that object (marked by entrainment, or re-using the same shortened referring expression) \citep{BrennanClark1996, GalatiBrennan2021}, as in this sequence of repeated references to the same object over multiple rounds \cite[adapted from][p. 1488]{BrennanClark1996}:

% \item
\vspace{0.5em}
\noindent
\textit{Round 1:} ``a car, sort of silvery purple colored''\\
\textit{Round 2:} ``purplish car going to the left'' \\
\indent . . . \\
\textit{{Round 5:}} ``the purple car''
\vspace{0.5em}

In another study that required triads of strangers to reach consensus while recalling the events from a movie clip that they had watched earlier, the speakers often hedged their contributions to the conversation, presumably to mark a lack of certainty about an utterance and an openness to being corrected by their partners \cite{BrennanOhaeri1999}. For example, from a triad that communicated by speaking face-to-face:
\vspace{-0.5em}
\begin{quote} \it
Yeah, they were sitting around the fireplace in the night... sort of like a bedtime story kind of thing
\end{quote}
\vspace{-0.5em}

\noindent
People who did the same task by texting rather than speaking used fewer words, but still hedged:
\vspace{-0.5em}
\begin{quote} \it
We all agree it was a wreathy thingy on his neck???
\end{quote}
\vspace{-1em}

\subsection{How Listeners React to Hedges} 

Hedges convey meaningful information that can affect listeners’ subsequent behavior; a handful of psychological studies have measured the impacts of hedges on listeners. For example, children exposed to new words from a speaker who hedged learned fewer novel words compared to children exposed to a speaker who did not hedge \cite{Sabbagh}. Listeners rated utterances as more uncertain when they included shield hedges (e.g., “I think it was a mug”), and these ratings were related to speakers’ ratings of their own uncertainty in identifying an image \cite{PogueTanen2018}. Moreover, addressees in a referential communication task expended more effort while grounding (they produced more low-confidence responses such as clarification questions) to demonstrate understanding when the speaker’s description had contained a hedge \cite{Dahan2023}. 

Hedges also influence which details are retold to another person; in one study, hedged details were less likely to be repeated to another addressee as compared to unhedged details \cite{FoxTree}, although in the same study, hedged information presented in a story was more likely to be remembered by listeners; this was thought to stem from deeper engagement with hedged information when it was first presented \cite{FoxTree}. And in tutoring dialogues, where face management can be particularly important, students were more successful at solving problems when their peer tutors used hedges \citep{Madaio2017IJCSCL}.

\begin{table*}[t]
\centering
\resizebox{\linewidth}{!}{
\begin{tabular}{l|l}
\toprule
\textbf{Hedge Type}                                                        & \textbf{Example(s)}                                     \\ \hline
Like (not used as a simile, verb, or comparison)                           & "and then he like went over by..."                           \\
You know (not to communicate another's knowledge or as a discourse marker) & "and you know as he's falling down"                     \\
Just (not used to mean "only")                                             & "he just jolts away"                                    \\
Approximators/Rounders                                                     & "kind of", "about"                                      \\
Proxies (for a detail the speaker cannot or chooses not to recall)         & "thing," "whatever," "or something," "and everything"   \\
Morpheme suffixes to content words                                         & "circley," "springy"                                    \\
Expressions of doubt attached to claims; self-speech                       & "I don't know," "maybe," "I guess," "what's it called?" \\
Tag questions and try markers                                                             & "he's standing there, right?"                         \\ \hline
\end{tabular}
}
\caption{Coding scheme used to mark hedges in corpus.}
\label{tab:hedgescoding}
\end{table*}

\section{Related Computational Work}
\label{sec:rel_work}
\subsection{Hedging}
%<Describe Hirschberg and Cassell papers>
Several research programs have examined hedges and the criteria for coding them, with computational goals that include automatic hedge detection.  Hedging is domain-specific, meaning that their forms and frequencies vary across corpora; they are also context-specific, as they cannot be identified accurately simply by searching for strings \citep{Prokofieva2014HedgingAS}. Hedges are distributed differently within different corpora (ibid).

Hedges are often ambiguous and difficult to code in the absence of dialogue context. In “I think it’s a little odd,” \textit{I think} is often a hedge, but might not be when proffered in response to a question (“So what do \textit{you} think?”). Hedges in spoken utterances may be disambiguated by stress and other intonational cues, as in “\textit{I} think he’ll win!” (not a hedge) vs. “I \textit{think} he’ll win?” (a hedge). 
%Hirschberg and colleagues
Previous work found many cases of tokens that can serve as hedges as well as non-hedges, with systematic tests for coders to use in annotating them for gold standards \citep{Prokofieva2014HedgingAS, Ulinski2019CrowdsourcedHT, ulinski-etal-2018-using}.

The coding of hedges is complicated by the fact that in spoken dialogue, they often co-occur with speech disfluencies. In some contexts, it may be difficult to distinguish these two kinds of signals \citep{Prokofieva2014HedgingAS}, particularly since listeners can use disfluencies in much the same way they can use hedges to draw conclusions about the speaker's mental state \citep{Arnold2003, Arnold2007}

A strong motivation for computational work on hedging comes from work on computer-assisted learning by Cassell and colleagues, specifically tutoring dialogues 
%are a particularly relevant domain in which to examine hedging 
\citep{abulimiti-etal-2023-generate, abulimiti2023kind, raphalen-etal-2022-might}. 
% \john{Added here about raphalen}
% \susan{Great! John, is my wordsmithing toward the end OK? @john} -> JM: looks good.
Most similar to our work is \citet{raphalen-etal-2022-might}, where the authors propose a model that combines rule-based classifiers and machine learning models with interpretable features such as unigram and bigram counts, part-of-speech tags, and LIWC categories to identify and classify hedge clauses. Our work differs in two major ways: first, our work operates on the \gls{token} level rather than on the clause level. Token level classification makes possible a truly end-to-end approach (classifying all hedge and non-hedge tokens in utterances). Second, we include experiments with modern LLMs and offer a detailed error analysis into their mistakes; stemming from this error analysis, we use an 
%novel\susan{NOT NOVEL! - see https://arxiv.org/pdf/2310.15100 (but  that has not been peer reviewed} 
LLM-in-the-Loop approach \cite{dai-etal-2023-llm} to correcting gold standard hedge codings. 

\subsection{Belief}

Hedging and the notion of belief (how committed the speaker is to the truth of an event) are closely related; hedges are often used by speakers to indicate a lack of belief or commitment towards what they say.  
% \susan{John and Owen, can we also cite your Interspeech paper here or add a bit at end of this paragraph? Note that we should try not to include the length} -> JM: Interspeech is still under review. I cited our other belief work in Machine Learning Approaches
\citet{ulinski-etal-2018-using} improved belief classification using a hedge detector, yielding an improvement for the non-committed and reported belief labels.

\paragraph{Corpora} Several corpora have been created that annotate the author's degree of belief \citep{diab-etal-2009-committed,prabhakaran-etal-2010-automatic,lee-etal-2015-event,stanovsky-etal-2017-integrating,rudinger-etal-2018-neural-models,pouran-ben-veyseh-etal-2019-graph,jiang-de-marneffe-2021-thinks}. There are two corpora that further annotate nested beliefs of the sources mentioned in the text: FactBank 
\citep{sauri2009factbank} and the Modal Dependency corpus \citep{yao-etal-2021-factuality}.

\paragraph{Machine Learning Approaches} Modern neural methods for belief detection include \gls{LSTM}s with multi-task or single-task approaches \citep{rudinger-etal-2018-neural-models}, using \gls{BERT} representations alongside a graph convolutional neural network \citep{pouran-ben-veyseh-etal-2019-graph}, or fine-tuning BERT with a span self-attention mechanism \citet{jiang-de-marneffe-2021-thinks}. Recent state-of-the-art work finds that fine-tuning RoBERTa \citep{murzaku-etal-2022-examining} or fine-tuning Flan-T5 \citep{murzaku-etal-2023-towards} yields the best performance on most corpora. For the label \textit{Underspecified} (or, corresponding to no commitment and/or a hedge), these modern methods yield f-measures in the low to high 80s. We also have prior work exploring multi-modal approaches to belief detection \citep{murzaku-et-al-2024-multimodal}.

\section{The Roadrunner-Hedge Corpus}
\label{sec:corpus}
For training and testing, we obtained a corpus \cite{GalatiBrennan2010} of spontaneous narratives produced from memory by 20 speakers who had watched a Roadrunner cartoon. Each speaker narrated the story face-to-face to an audience, a total of three times: first to a naïve addressee, a second time to the same addressee, and a third time to a new naïve addressee (with the latter two episodes counterbalanced for order). The original experiment was designed to detect differences in collateral signals (intelligibility vs. attenuation of speech and gestures) stemming from the speaker’s vs. the addressee’s knowledge states--that is, whether the story was new for the speaker (told for the first time) vs. old (retold), compared to the addressee’s knowledge state (new vs. heard for the second time). Findings included that the attenuation of both referring expressions \citep{GalatiBrennan2010} and gestures \citep{Galati2014SpeakersAG} were driven by \textit{both} speakers’ and addressees’ knowledge states--that is, shortened upon retelling the story to the same addressee, but lengthened upon retelling to a new addressee.

\paragraph{Gold Standard Coding.} The original corpus transcribed the spontaneous narratives in detail, including speaking turns and disfluencies \cite[for details, see][]{GalatiBrennan2010}, segmented into lines by installments that corresponded to \glspl{narrative element}s in the cartoons. We annotated hedges on the original Roadrunner corpus to create the gold standard for hedge training and detection (the \textit{Roadrunner-Hedge} corpus; see \url{\repo}
%\href{\repo} 
for the annotation codebook).  

The Roadrunner-Hedge corpus is distributed as a csv file.  It is structured as a total of 5,508 lines, over a quarter of which (N=1424) include one or more hedges. The first author annotated hedges in the corpus as in Table~\ref{tab:hedgescoding}. Although disfluencies such as fillers (\textit{uh, um}) and re-starts can function as hedges, we made a principled decision to not code them as such; hedges in our corpus are presumed to be shaped by the speaker's intention, whereas disfluencies are not necessarily under a speaker's control as a communicative signal, but may reflect difficulties in speaking \cite{grice1957meaning, clark1994}.
%modifiable according to a speakers' intentions. See][for discussion]{Brennanetal2010}.
%\amie{added language about disfluencies; this was an in-principle choice and not case-by-case.}
Overall word counts for hedges and non-hedges are 1,728 and 38,018 words respectively. Most hedges are one word, but a few cases contain many words. For each line in the csv file (corresponding to a narrative element), hedges are listed (separated by commas) in an adjacent cell. %Overall, there were 1803 annotated hedges, 
Each line has an average of 0.33 hedges. 

\paragraph{Inter-Rater Reliability.} 

To compute inter-rater reliability, a trained research assistant coded 7 randomly-selected transcripts with no overlapping speakers (10\% of the corpus). 
%The transcript was segmented by breath group in Excel, with hedges listed (separated by commas) in an adjacent Excel cell. 
We calculated \gls{Cohen's Kappa} from each word marked as a hedge within each transcript. There was high agreement between coders, with $\kappa=0.985$.
%\amie{updated this to show IAA at the WORD level, per Owen's comments.}
%\amie{After we get to the post-review stage, comment on using 21 triads vs the original paper using 20}

\paragraph{Corpus Analysis.} The Roadrunner-Hedge corpus, like the tutoring dialogues used by \citet{abulimiti2023kind, raphalen-etal-2022-might}, has fewer cases with hedges than without, but with more hedges per segment overall (25.85\% of lines vs. 14.26\% of turns respectively).

Over the three versions of the cartoon story produced by each speaker, hedges were most frequent in the first telling when the story was new to both speaker and addressee and least frequent when told to the same addressee a second time, consistent with the original findings from \citeauthor{GalatiBrennan2010} that collateral signals are affected by the knowledge states of both speaker and addressee.

%\begin{table}[htb]
%\centering
%\begin{tabular}{|c|c|}
%\hline
 %& Count \\ \hline
%Hedges & 1,728 \\ \hline
%Non-hedges & 38,018 \\ \hline
%\end{tabular}
%\caption{Token count for hedges and non-hedges in the annotated Roadrunner corpus. }
%\label{tab:label_counts}
%\end{table}

\begin{table*}[ht]
\centering
\begin{tabular}{lll|ccc}
\toprule
\textbf{Model} & \textbf{Training} & \textbf{Prompt} & \textbf{Precision (P)} & \textbf{Recall (R)} & \textbf{F1 Score (F1)} \\
\midrule
BERT & Finetuned & - & $0.883 \pm 0.015$ & $0.934 \pm 0.012$ & $0.908 \pm 0.010$ \\
GPT-4o & Few-Shot & List & $0.613 \pm 0.027$ & $0.848 \pm 0.018$ & $0.712 \pm 0.021$ \\
LLaMA-3 & Few-Shot & List & $0.518 \pm 0.035$ & $0.799 \pm 0.022$ & $0.628 \pm 0.031$ \\
GPT-4o & Few-Shot & BIO & $0.514 \pm 0.024$ & $0.766 \pm 0.036$ & $0.616 \pm 0.030$ \\
GPT-4o & Zero-Shot & List & $0.430 \pm 0.014$ & $0.711 \pm 0.004$ & $0.536 \pm 0.012$ \\
GPT-4o & Zero-Shot & BIO & $0.436 \pm 0.026$ & $0.618 \pm 0.033$ & $0.510 \pm 0.028$ \\
LLaMA-3 & Few-Shot & BIO & $0.298 \pm 0.018$ & $0.625 \pm 0.016$ & $0.404 \pm 0.019$ \\
LLaMA-3 & Zero-Shot & BIO & $0.167 \pm 0.014$ & $0.428 \pm 0.019$ & $0.240 \pm 0.017$ \\
LLaMA-3 & Zero-Shot & List & $0.274 \pm 0.023$ & $0.146 \pm 0.010$ & $0.190 \pm 0.011$ \\
\bottomrule
\end{tabular}
\caption{Average performance metrics over the five folds with standard deviations for different models, training methods, and prompt types, ordered by F1 score.}
\label{tab:performance_metrics_ordered}
\end{table*}

\section{Experiments}
\label{sec:experiments}
%\john{5/20: Addressed Owen's comments for this whole section. Please reread and delete this coment when approved.}
%\amie{re: John's comment above, this looks good to Amie and @Susan}
\subsection{Experimental Setup}
In this section, we present our hedge classification experiments on the Roadrunner-Hedge corpus, 
%described in Section~\ref{sec:corpus}
conducted by fine-tuning BERT and performing zero-shot and few-shot experiments with state-of-the-art LLMs. %{tab:label_counts}
%\adil{table no longer exists}
For all experiments, we performed five-fold cross validation using a fixed seed (42), splitting the corpus into a 80/20 train/test split. For our fine-tuning experiments, we did not perform any hyperparameter tuning, and therefore do not have a validation set. 

We performed all zero-shot, few-shot, and fine-tuning experiments on the fold's respective test sets and report the average and standard deviation over all five folds test sets for \gls{F1}, \gls{precision}, and \gls{recall}.
\subsection{Zero Shot and Few Shot}
\label{subsec:zfs}
For the zero-shot and few-shot experiments, we used GPT-4o \citep{gpt4o} and LLaMA-3-8B-Instruct \citep{llama3modelcard}, as these two LLMs have achieved state-of-the-art results in many zero-shot or few-shot benchmark tasks. 

We conducted two classes of zero-shot and few-shot experiments: count/list generation and \gls{BIO} tag generation. Both prompts began with an instruction detailing the specific task, and a random example. In our few-shot experiments, we provided three fixed hand-crafted examples. For our count/list generation, we prompted the models to list the integer number of hedges present in the utterance and then generated a list of the exact hedge words. For our BIO tag generation, we generated the tokens and their respective tags, where label \textit{B} represents the beginning of a hedge token or span, \textit{I} represents the inside of a hedge span, and \textit{O} represents another token, all separated by \textit{``/''}. For example, given the utterance \textit{It is like warm}, we prompted the model to generate \textit{It/O is/O like/B warm/O}.

 We provide our exact prompts with their corresponding instructions in Appendix~\ref{app:experiment-details}. For our GPT-4o experiments, we used the default OpenAI API hyperparameters and a \gls{temperature} of 1.0. 

\subsection{Fine-tuning}
We performed all fine-tuning experiments using BERT \citep{devlin-etal-2019-bert}, specifically bert-base-uncased. We also performed experiments with the large variants of the model (bert-large), newer encoder-only models like RoBERTa \citep{liu2019roberta} and DeBERTa-v3 \citep{he2021debertav3}, and encoder-decoder models like Flan-T5 \cite{chung2022h}, but got either worse or closely similar results.

\paragraph{Task Description} All experiments followed a standard BIO token labelling approach to classify hedge tokens (B), tokens inside of hedge spans (I), and all other tokens (O). In other words, given an input utterance of n tokens, the respective BIO labels were output for each of the n tokens. Following the same example as described in our zero-shot and few-shot experiments in Section~\ref{subsec:zfs}, we fine-tuned BERT to classify the tokens as {\em It/O is/O like/B warm/O}.

\paragraph{Hyperparameters} We followed a standard fine-tuning approach, fine-tuning for a fixed 5 \gls{epoch}s. We set the batch size to 16 and learning rate to 2e-5. We performed five-fold cross validation and test on each folds respective test set. We did not perform any hyperparameter tuning. 
\subsection{Results}
The performance of the models is shown in Table \ref{tab:performance_metrics_ordered}, which reports average precision (P), recall (R), and F1 over the five-folds. For our zero-shot, few-shot, and fine-tuning experiments, these metrics are calculated on each fold's test set and then averaged.

% \paragraph{Fine-tuning vs. Prompting} 
Despite its much smaller parameter count, BERT fine-tuned for BIO tagging outperforms even the best scoring prompting approaches by nearly 20 points in F-measure. This is consistent with a general trend in the literature of more parameter efficient fine-tuning approaches outperforming larger zero-shot and few-shot methods \citep{liu-etal-2022-few}, though the gap here is larger than one might expect.

% \paragraph{Zero-Shot vs. Few-Shot}
% Analysis of this here.
% https://colab.research.google.com/drive/1k34jZ9G6D0dpH2wtPy-fGwalLFD5YzU_?usp=sharing
In comparisons of the zero-shot and few-shot prompting methods, the few-shot models, unsurprisingly, performed better. The few-shot experiments averaged an F1 of 0.59, 22 points higher than the zero-shot models average of 0.37.

% \paragraph{Listing vs. BIO Tagging}
Of the two output formats prompted for, listing and BIO, the listing approach performed better. On average, models instructed to output a list had an F1 of 0.52 compared to 0.44 for those instructed to perform BIO tagging. 
% is BIO more typical for this type of task? Is it surprising that listing performed better?

Among the two LLMs prompted, GPT-4o always performed best. Across all models and approaches, including fine-tuned BERT, precision tended to be lower than recall, with a mean of 0.46 for precision compared to 0.65 for recall. In other words, the models over-predicted the presence of hedges.

\begin{table*}[ht]
\centering
\begin{tabular}{llllllll}
\textbf{Gold Errors} &    & \textbf{False Negative} &    & \textbf{False Positive} &    & \textbf{Other}         &    \\ \hline
\textit{Like}               & 13 & Approximator            & 9  & \textit{Like}           & 13 & \textit{I} should be \textit{B} & 4  \\
Proxy                & 12 & Proxy                   & 8  & \textit{Just}           & 8  & \textit{O} should be \textit{I} & 3  \\
\textit{Just}                 & 7  & Self-talk               & 4  & False proxy    & 4  & \textit{B} should be \textit{I} & 2  \\
Approximator         & 1  & \textit{Like}                    & 3  & \textit{You know}       & 2  & Other         & 2  \\
Other                & 1  & \textit{Just}                    & 1  & Misc. word       & 2  &               &    \\
                     &    & Morpheme                & 1  &                &    &               &    \\ \hline
\textbf{Total}       & \textbf{34} &                         & \textbf{26} &                & \textbf{29} &               & \textbf{11}
\end{tabular}
\caption{Expanded error analysis on the BERT fine-tuned model, by hedge type.}
\label{tab:error2}
\end{table*}
\begin{table*}[ht!]
\centering
\begin{tabular}{llllllll}
\textbf{Gold Errors} &            & \textbf{False Negative} &             & \textbf{False Positive} & \textbf{}   & \textbf{Other} &            \\ \hline
Approximator         & 4          & \textit{Just}                    & 12          & Disfluency tag          & 37          & Other          & 1          \\
\textit{Just}                 & 1          & Proxy                   & 8           & Misc. word                & 15          &                &            \\
\textit{Like}                 & 1          & \textit{Like}                    & 3           & \textit{Like}                    & 7           &                &            \\
Proxy                & 1          & Morpheme                & 1           & Approximator            & 3           &                &            \\
Self-talk            & 1          & Self-talk               & 1           & Intensifiers        & 3           &                &            \\
                     &            &                         &             & \textit{You know}                & 1           &                &            \\ \hline
\textbf{Total}       & \textbf{8} & \textbf{}               & \textbf{25} & \textbf{}               & \textbf{66} & \textbf{}      & \textbf{1}
\end{tabular}
\caption{Expanded error analysis on the GPT-4o FSL model, by hedge type.}
\label{tab:error3}
\end{table*}

\section{Error Analysis}
\label{sec:ea}
%\amie{updates for this entire section, delete this comment when approved}
While the fine-tuned BERT model performed fairly well, a certain number of cases did not align with the gold labels in the data. We performed error analysis to understand whether there were any systematic deviations from the corpus annotation.

We conducted an error analysis on the top two performing models, the fine-tuned BERT model and the GPT-4o Few-shot List (FSL) model (F1 = 0.91 and 0.71, respectively). Starting with the first fold, we selected the first hundred errors to categorize. These errors are broadly divided into instances where the models failed to detect a hedge (false negatives) and instances where models returned cases that were not annotated hedges (false positives). The remaining errors fell into two other categories: a gold error category, wherein errors in the (human) annotation were discovered, and an ``other'' category. 

Of the hundred errors sampled from the BERT model, approximately the same number of errors were false negatives (26) as false positives (29). Of the hundred errors sampled from the GPT-4o FSL model, 66 were false positives and 25 were false negatives (reflecting the low precision and higher recall for this approach; see Table~\ref{tab:error2} and~\ref{tab:error3} for full error descriptions for BERT and GPT-4o FSL models). 

Although the corpus annotation does not include the \textit{type} of hedge (only the presence or absence of hedge tokens), our error analysis looked at hedge types in order to tease apart model behaviors. We observed systematic differences between models in their types of mismatches with the gold standard.

\textbf{False Positives.} First, the GPT-4o FSL model inaccurately classified disfluencies (e.g., ``uh'') as hedges in 37 of the 66 false positives reviewed, whereas BERT did not. Second, BERT showed quite a different pattern of mismatches than GPT-4o when classifying ``like'', returning false positives that always turned out to be comparatives (e.g., ``it's like an open elevator''). These we considered to be true errors in their text form, although some may be ambiguities that could be resolved prosodically.
%BERT was likely overexposed to "like" hedges cases in training, given that 978 of the 1065 tokens of "like" in the whole corpus were originally annotated as hedges by the gold standard.

\textbf{False Negatives.} Tokens denoting approximator hedges (e.g. ``that's \textit{basically} it'') were frequently misclassified as false negatives by BERT (9 of 26 false negatives reviewed), but never by the GPT-4o FSL model.

In addition, \textbf{Other} emerged as a category type for situations that could not clearly be described as false positives, false negatives, or gold errors. In the BERT model, these cases were typically segmentation errors (i.e., an inner token mislabeled as a beginning token). 

Notably, the largest class of errors for the BERT model was the \textbf{Gold Error} category (34 of 100). This was not the case for the GPT-4o model (only 9 gold errors). The BERT fine-tuned model revealed mistakes made by the human annotators for hedges denoted by ``like'', ``just'', and proxy hedges (e.g. ``and stuff''). Upon closer inspection, some of these cases were ambiguous. For example, ``he just hits the ground'' could be taken to mean that the only action performed was hitting the ground (where ``just'' means only) or ``just'' might function to reduce the speakers' certainty \cite[as in][]{Madaio2017IJCSCL}. Again, the text format of the storytelling corpus leaves some interpretations ambiguous that could be clarified with signals such as timing and prosodic stress.%\susan{perhaps refer to the potential for prosody to disambiguate this in future work -- hedges may have different stress patterns than non-hedges}
%\amie{was thinking the same thing}

%Additionally, because we exclude instances of ``like" that compare (see Table ~\ref{tab:hedgescoding}), we include instances of like that demonstrate \cite{Clark1990QuotationsAD}. 

% this table was removed
%\susan{I will beef up this paragraph with LLM-in-the-Loop}
The number of Gold Errors identified by the BERT model allowed us to modify the original gold annotation with missed cases and to re-evaluate the performance of our models more accurately -- a sort of \textit{LLM-in-the-Loop} approach (see Table~\ref{tab:er}). 
%PUT THIS ELSEWHERE, AFTER WE GET REVIEWS BACK The corrected corpus will be made public.

\begin{table*}[ht!]
    \centering
    \begin{tabular}{lccc}
        \toprule
        \textbf{Model} & \textbf{Original Gold F1    } & \textbf{LLM-in-the-Loop Gold F1   } & \textbf{Error Reduction (\%)} \\
        \midrule
        BERT & $0.908 \pm 0.010$ & $0.925 \pm 0.019$ & 18.5\% \\
        GPT-4o Few-Shot & $0.712 \pm 0.021$ & $0.721 \pm 0.020$ & 3.1\% \\
        GPT-4o Zero-Shot & $0.510 \pm 0.028$ & $0.551 \pm 0.011$ & 8.4\% \\
        \bottomrule
    \end{tabular}
    \caption{F1 scores with standard deviations on the original corpus, F1 scores with standard deviations obtained on the corpus corrected after LLM-in-the-Loop, and the change in average performance for our top performing models.}
    \label{tab:er}
\end{table*}

\section{Discussion}
\label{sec:discussion}

The results show that even enormous, recently released LLMs cannot reliably recognize hedges.  There is no ``emergent'' ability in LLMs to understand full human linguistic behavior.  On the other hand, when we explicitly train a small, rather old LLM (BERT) to perform our task by fine-tuning it, it performs quite well.  What this shows is that detecting hedges is a capability that can be learned, but it cannot be learned in the manner that LLMs are taught, namely by simply ingesting large amounts of varied data.  We interpret this to mean that if we want to make LLMs able to converse with humans as humans do, we need to understand what capabilities LLMs need and how to provide them with the ability to do so.

%It is perhaps surprising at first that the best-performing model is BERT, since BERT is the oldest and smallest model. However, we do note that BERT introduced the classification token ([CLS]), allowing for straightforward and powerful text classification abilities with a classification head \citep{devlin-etal-2019-bert}. Furthermore, BERT was pre-trained for masked language modelling (MLM), making it more suitable than generative LLMs for token classification tasks such as ours. Our few-shot GPT-4o results, while much lower than BERT's results, still prove competitive. 

The prevalence of gold errors discovered by the BERT model raises two interesting points for discussion. First, some of these discrepancies identified by the BERT model were clearly errors made by the human coders; this was true in particular for proxies, which BERT coded for hedges more consistently than did human coders. This error analysis allowed us to iteratively improve the human coding before the final analysis, essentially deploying an LLM-in-the-Loop approach. 
Second, the discrepancies between BERT and gold coding on the tokens \textit{just} and \textit{like} highlight that these types of hedges have high potential for ambiguity--perhaps the very sort of ambiguity that could be resolved by prosody. 

%Be sure to discus the case of like in this particular corpus -- lots of demonstrations due to the nature of the physical comedy of RR cartoons!

%\susan{why Bert performed best. Is it relevant that collateral signals such as hedges differ between written text and spontaneous speech?}
%\adil{is it relevant that BERT is the only autoencoding model tested?}
%\susan{Adil, pls expand on this!}

\section{Limitations and Future Work}
\label{sec:fw}
This work represents the first step in our research program that aims to train LLMs to use collateral signals in support of human-LLM dialogue. Once hedges can be recognized by an LLM, it remains to be shown that they can be meaningfully interpreted and generated. Relevant work by Cassell and colleagues has shown that it is possible to generate hedges in tutoring dialogues, but not always positioned where they are most probable or useful \citep{abulimiti-etal-2023-generate}. In future work, we plan experiments using top-performing models such as BERT and GPT-4o in high- and low-probability situations that systematically vary the certainty associated with prompted-for information (where hedges can be most useful). It is already clear from our pilot trials using ChatGPT 3.5 that LLMs hedge somewhat superficially (hedging where humans wouldn't and failing to hedge where humans would).

\paragraph{Domains of Dialogue.} Here we have used human-generated dialogue from a single domain, retelling stories from Roadrunner cartoons; the training data are text transcripts of speech. Because the initiative was unbalanced in this collaborative task, most of the speaking in each triad was done by the the partner who viewed and retold the cartoon stories in series to the two co-present addressees. 

A more balanced domain in which partners continuously monitor each other's understanding to do a physical task--such as matching pictures of difficult-to-describe objects--could yield more hedges, distributed differently. We plan to conduct similar tests to replicate the current results on such referential communication corpora collected previously in our lab.

It is interesting that despite the fact that there is not a single instance of dialogue in Roadrunner cartoons (apart from Roadrunner's smug, trademark ``meep meep'' upon escaping from Coyote), speakers who retell the story in a dramatic and humorous way do a great deal of what looks like quoting Coyote's and Roadrunner's reactions:
\vspace{-0.5em}
\begin{quote} \it
so then he’s saying he’s like gone all sad and stuff you know?
\end{quote}
\vspace{-0.5em}
% and
\vspace{-0.5em}
\begin{quote} \it
and he’s like whatever she’s gonna be dead right?
\end{quote}
\vspace{-0.5em}
Such uses of \textit{like} in this corpus match the quotation-as-demonstrations forms described by \citet{Clark1990QuotationsAD}; they count as hedges in that the speaker marks what follows as \textit{not verbatim}.

%JM: Began writing audio section. @Susan @Amie please edit if I miss anything or said sometrhing wrong

\paragraph{Training with audio input.} Our results for detecting hedges in this transcribed spoken corpus are surprisingly strong, especially given that the LLMs we used were pre-trained primarily on originally written text. But it is well-known that features such as pausing and intonation are related to speakers' levels of commitment to and confidence in their utterances. We plan to incorporate audio into future hedging studies and will explore multi-modal neural architectures fusing both speech and lexical features as we did in \citep{murzaku-et-al-2024-multimodal} for belief recognition.

%\paragraph{Video} On top of text transcripts and audio files, the corpus also contain recorded videos of the interactions \john{Left this half written, @Susan or @Amie i believe tou can write this part better than me} \susan{John, a third or more of the videos could not be salvaged when I tried to have them professionally restored, so we will not be sharing the audio and video files.}

\paragraph {Reliability.} It is critical to keep in mind that human and LLMs are very different sorts of agents. Psychometric tests show that individual humans are likely to respond consistently when tested repeatedly, whereas an LLM is not \cite{shu2024dont}. LLMs have no sense of ``self'' and are likely to respond differently when re-prompted with the same prompt. To the extent that a hedge signals that a speaker does not wish to be held entirely accountable for what they're saying, hedging on the part of an LLM may actually be desirable as a way to encourage users to not assume they can hold it accountable. On the other hand, it may be desirable for an LLM to be able to signal its \textit{confidence} -- the reliability or quality (or lack thereof) of information it's presenting -- through the presence or absence of hedges. Finally, it remains to be seen whether LLMs can learn about interaction through exposure to collateral signals in meaningful contexts.

%TO DO:
%Acknowledgments:
% DARPA CCU, AI cluster, Seawulf, IACS in general (?), BIAS-NSF
 
\section{Conclusion}
\label{sec:conclusion}
Our project is grounded in psycholinguistic theory and aims to capture theory-of-mind aspects of hedging among discourse participants. We present the Roadrunner-Hedge corpus, with hedges annotated from naturally occurring dialogues by speakers describing Roadrunner cartoons. We use the corpus to train and perform experiments on detecting hedges using BERT, GPT-4o, and LLaMA-3. We find that fine-tuning BERT significantly outperforms state-of-the-art LLMs in few-shot and zero-shot settings. With our systems outputs, we perform an error analysis and use an LLM-in-the-Loop approach to correct gold standard annotations. Our LLM-in-the-loop approach provided further error reductions on all models.

\section*{Ethical Considerations}
The Roadrunner-Hedge corpus was collected with Institutional Review Board approval from undergraduate students who gave informed consent prior to participating in the experiments% conducted in \citep{GalatiBrennan2010, Galati2014SpeakersAG}
.

\section*{Acknowledgments}
This material is based upon work supported in part by the National Science Foundation (NSF) under No. 2125295 (NRT-HDR: Detecting and Addressing Bias in Data, Humans, and Institutions) as well as by funding from the Defense Advanced Research Projects Agency (DARPA) under the CCU program (No. HR001120C0037, PR No. HR0011154158, No. HR001122C0034).  Any opinions, findings and conclusions or recommendations expressed in this material are those of the author(s) and do not necessarily reflect the views of the NSF or DARPA.

We thank both the Institute for Advanced Computational Science and the Institute for AI-Driven Discovery and Innovation at Stony Brook for access to the computing resources needed for this work. These resources were made possible by NSF grant No. 1531492 (SeaWulf HPC cluster maintained by Research Computing and Cyberinfrastructure) and NSF grant No. 1919752 (Major Research Infrastructure program), respectively.

We would also like to thank our reviewers for their helpful comments, as well as Kayla Hunt for assistance with reliability coding. 
% This material is based upon work supported by the National Science Foundation under Award No. 2125295 (NRT-HDR: Detecting and Addressing Bias in Data, Humans, and Institutions) and DARPA CCU.
\bibliography{anthology,latex/custom}

\appendix
%\section{Corpus Details}
%The corpus of spontaneously told stories were generated  

\section{Prompting Details}
\label{app:experiment-details}

The exact prompt templates used for the BIO and listing experiments are shown below.

% \begin{figure*}[htp]
\begin{lstlisting}
Given an utterance, perform BIO tagging to classify hedges in the sentence. ``BIO" tagging is a method used in named entity recognition where each token (word) in the sentence is tagged as follows:

B (Beginning): The token is the beginning of a hedge.
I (Inside): The token is inside, but not the first token of a hedge.
O (Outside): The token is not part of a hedge.
Please assign one of these tags to each token in the given utterance, representing whether each word is part of a hedge phrase or not. Format your response by listing each token followed by its corresponding BIO tag.

Example:

If the utterance is ``I think maybe you could try an approach like that" then ``I think" and ``maybe" are identified as hedges so your output should look like this:

Utterance:
I think maybe you could try an approach like that

Tags:
I/B think/I maybe/B you/O could/O try/O an/O approach/O like/O that/O

Now given the following input, please classify the hedges in the sentence.

Utterance:
{utterance}
\end{lstlisting}

\begin{lstlisting}
Given a conversation, answer a question. Be as precise and succinct as possible. If asked for a number, provide a numeric value.

Format the output as follows:
Number of Hedges: Integer number of linguistic hedges (e.g. 0)
List of Hedges: List of hedges found (e.g. [``first hedge", ``second hedge", etc...])

Conversation:
{utterance} <stop sign emoji>

Question:
At the line that ends with <stop sign emoji>, how many linguistic hedges are there? List all the linguistic hedges using quotations. Do not add any additional information.
\end{lstlisting}
% \end{figure*}

% \section{Experimental Details}
% \paragraph{LLaMA-3 Hyperparameters} We use a temperature of 1.0 and the default generation hyperparameters in HuggingFace transformers.

% \paragraph{GPT-4o Hyperparameters} We perform all experiments using the OpenAI API. We use a temperature of 1.0 and the default generation hyperparameters. 

% \paragraph{Packages}To fine-tune BERT and perform zero and few-shot experiments with LLaMA-3, we used the transformers library provided by HuggingFace \citep{wolf2019huggingface}. All code for experiments and pre-processing will be made available.

\section{Glossary}
Due to the interdisciplinary nature of this work, we provide below brief definitions for terms which may be unfamiliar.  
The numbers refer to the pages in this paper in which the term first appears.

\printnoidxglossary

\end{document}